\documentclass{article}
\usepackage{spconf,amsmath,graphicx,hyperref}

\usepackage{pdfpages}
\usepackage{enumitem}
\usepackage{booktabs}
\usepackage{multirow}

\usepackage[normalem]{ulem}
\useunder{\uline}{\ul}{}
\usepackage{float}  % 放在导言区
\usepackage{setspace} % 放在导言区
\usepackage{subcaption}

\title{DIRECT: Direct Decoding for Efficient and Aligned Sequence Labeling with Large Language Models}
\name{Yilei Wang\textsuperscript{1}, Jiaxin Gan\textsuperscript{1}, Kexuan Zhang\textsuperscript{1}, Ling Li\textsuperscript{1}, Wentao Zhang\textsuperscript{2}, Peichao Lai\textsuperscript{2, *}\thanks{* Corresponding author: \href{mailto: lpc@pku.edu.cn}{lpc@pku.edu.cn}}}
\address{\textsuperscript{1}College of Computer and Data Science, Fuzhou University, Fuzhou, 350108, China \\
\textsuperscript{2}School of Computer Science, Peking University, Beijing, 100871, China}
\begin{document}
\ninept
\maketitle
\begin{abstract}
Sequence labeling is a fine-grained information extraction task, yet existing large language model-based approaches suffer from insufficient domain alignment and low inference efficiency. To address these issues, we propose DIRECT, a framework that addresses these issues through training-time optimization and inference-time rectification. Specifically, DIRECT performs Direct Preference Optimization (DPO) after supervised fine-tuning to strengthen task alignment with human preferences, and introduces a controlled decoding process that enforces fixed output formats and restricts predictions to candidate sets. To further improve efficiency, a template-filling mechanism requires the model to generate only label tokens while reusing prefixed content through the KV Cache, thus reducing redundant computation. Experimental results on eight datasets demonstrate that DIRECT achieves significant improvements in both performance and efficiency compared to existing methods.
\end{abstract}
\section{Introduction}
\label{sec:intro}

Sequence labeling is a fine-grained information extraction (IE) task that covers sub-tasks such as named entity recognition (NER), word segmentation, and part-of-speech (POS) tagging. It is crucial for other downstream natural language processing (NLP) tasks, such as co-reference resolution \cite{clark2016improving}, knowledge graph \cite{ji2021survey}, and question generation \cite{pergola2021boosting}.

Currently, large language models (LLMs), endowed with extensive knowledge bases and powerful language modeling capabilities, have been widely applied across a broad range of tasks \cite{ouyang2022training, wei2022chain,yang2024graphusion} . In the context of IE under low-resource scenarios, recent studies have primarily focused on fine-tuning open-source LLMs to enhance their performance, which can be broadly categorized into two approaches: (1) Multi-task generative frameworks: for example, UIE \cite{lu2022unified} resolves the challenge of heterogeneous task formats and limited transferability by unifying all IE tasks into a text-to-structure generative framework; USM \cite{lou2023universal} addresses the limitations of generative unification in terms of discriminative accuracy and cross-task sharing by reformulating IE as a semantic matching problem; and InstructUIE \cite{wang2023instructuie} further eliminates the need for task-specific fine-tuning in UIE by introducing an integrated framework for universal IE. (2) Auxiliary information-driven fine-tuning: for example, GNER \cite{ding2024rethinking} alleviates boundary ambiguity in generative NER by introducing negative instances during training; GoLLIE \cite{sainz2023gollie} improves model consistency in zero-shot IE by fine-tuning LLMs to explicitly follow detailed annotation guidelines; and C-ICL \cite{mo2024c} enhances few-shot learning performance by incorporating both correct examples and constructed hard negatives into in-context demonstrations for contrastive learning.

Despite the progress achieved by the aforementioned methods, several challenges remain: \textbf{(1) Insufficient domain-specific alignment.} In strict sequence labeling tasks, existing approaches rely solely on specifying task formats and output structures through instructions during fine-tuning, which cannot guarantee that the model outputs strictly align with the expected results. \textbf{(2) Low inference efficiency.} Most existing studies focus on training and fine-tuning LLMs, while still adopting the original autoregressive process during inference, leading to inefficiency in model decoding.

\begin{figure*}[ht]%
\centering
\includegraphics[width=\textwidth]{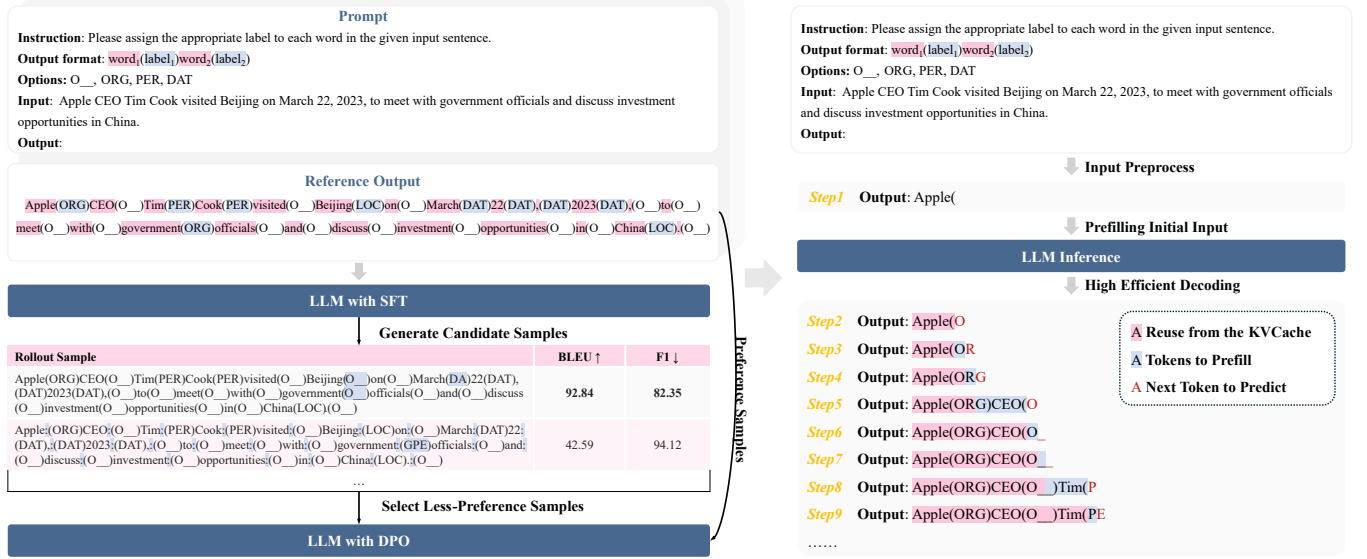}
\caption{The overall architecture of DIRECT.}\label{fig1}
\end{figure*}

To address the aforementioned challenges, we propose an innovative approach named \textbf{D}PO-based \textbf{I}nference \textbf{RECT}ification (\textbf{DIRECT}). \textbf{\textit{To enhance the domain-specific alignment of LLMs,}} we perform Direct Preference Optimization (DPO) training after supervised fine-tuning (SFT), thereby improving the model’s understanding of sequence labeling tasks and aligning its behavior more closely with human preferences. In addition, during inference, we intervene in the generation process by enforcing a fixed output format and restricting the generated content to a predefined candidate set. This dual mechanism of training and inference substantially improves the alignment of LLMs in domain-specific tasks. \textbf{\textit{To improve inference efficiency}}, we introduce a template-filling mechanism in which the model generates only the label components to be predicted, while the remaining parts are prefilled through the template with full reuse of the KV Cache. This approach avoids redundant generation and significantly enhances decoding efficiency.

Our contributions can be summarized as follows:

(1) \uline{\textit{Domain-adaptive alignment optimization.}} By incorporating DPO during the training phase and introducing controlled interventions in the generation process of LLMs during inference, our approach enables LLMs to better understand sequence labeling tasks and align more effectively with human preferences. (2) \uline{\textit{Significant improvement in inference efficiency.}} A template-filling mechanism requires the model to generate only label tokens, while the remaining parts are completed by the template with full KV Cache reuse, yielding up to \textbf{9$\times$ faster inference} compared with state-of-the-art methods. (3)\uline{\textit{State-of-the-art performance.}} We conduct extensive experiments on NER and POS tasks, and the results demonstrate that DIRECT achieves superior performance in sequence labeling tasks under low-resource settings.

\section{METHODOLOGY}
\label{sec:method}

\subsection{Task Definition}

Given a token sequence of length $n$, $X = \{x_1, x_2, \dots, x_n\}$, the objective of sequence labeling is to map it into a label sequence of the same length, $Y = \{y_1, y_2, \dots, y_n\}$. In this study, we formulate both NER and POS tagging as token-level classification tasks, where the expected model output is expressed as $O = \{x_1(y_1), x_2(y_2), \cdots, x_n(y_n)\}$. Figure~\ref{fig1} illustrates the overall framework of our proposed DIRECT method.

\begin{table*}[]
\caption{Overall performance evaluated using F1-scores, with \textbf{bold} values indicating the best results and \underline{underlined} values representing the second-best. The models in parentheses denote different backbones.}\label{tab1}
\centering
\begin{subtable}{\textwidth}
\resizebox{\textwidth}{!}{%
\begin{tabular}{@{}lcccccccccccc@{}}
\toprule
\multirow{2}{*}{\textbf{Dataset/Model}} & \multicolumn{3}{c}{\textbf{Youku}}               & \multicolumn{3}{c}{\textbf{Taobao}}              & \multicolumn{3}{c}{\textbf{Weibo}}               & \multicolumn{3}{c}{\textbf{Resume}}              \\ \cmidrule(l){2-13} 
                                        & $\mathcal{K}$=250          & $\mathcal{K}$=500          & $\mathcal{K}$=1000          & $\mathcal{K}$=250          & $\mathcal{K}$=500          & $\mathcal{K}$=1000          & $\mathcal{K}$=250          & $\mathcal{K}$=500          & $\mathcal{K}$=1000          & $\mathcal{K}$=250          & $\mathcal{K}$=500          & $\mathcal{K}$=1000          \\ \midrule
InstructUIE                             & 72.22          & \underline{72.83}    & \underline{78.03}    & 58.55          & 70.08          & 75.98          & 55.46          & \underline{66.30}    & 69.94          & 78.92          & 84.04          & 85.65          \\
GoLLIE                                  & \underline{72.34}    & 72.64          & 76.64          & 61.54          & 65.27          & 70.06          & 54.38          & 56.46          & 61.07          & 92.08          & 93.14          & 93.98          \\
GNER                                    & 60.55          & 65.38          & 72.97          & 44.24          & 55.88          & 64.28          & 35.73          & 48.19          & 59.00          & 82.23          & 87.84          & 90.98          \\ \midrule
\textbf{DIRECT(LLaMA-3.1-8B-Instruct)}   & 67.93          & 70.92          & 75.72          & \underline{65.79}    & \underline{72.68}    & \textbf{78.68}    & \underline{56.38}    & 63.71          & \textbf{70.52}    & \underline{93.07}    & \underline{93.30}    & \underline{95.13}    \\
\textbf{DIRECT(GLM4-9B-Chat)}            & \textbf{72.87} & \textbf{73.96} & \textbf{79.31} & \textbf{69.51} & \textbf{75.06} & \underline{78.58} & \textbf{64.14} & \textbf{69.79} & \underline{70.31} & \textbf{95.21} & \textbf{95.70} & \textbf{95.94} \\ \bottomrule
\end{tabular}
}
\end{subtable}

\vspace{0.2em} % 控制上下间距

\begin{subtable}{\textwidth}
\resizebox{\textwidth}{!}{%
\begin{tabular}{@{}lcccccccccccc@{}}
\toprule
\multirow{2}{*}{\textbf{Dataset/Model}} & \multicolumn{3}{c}{\textbf{CoNLL03}}             & \multicolumn{3}{c}{\textbf{MIT-Movie}}           & \multicolumn{3}{c}{\textbf{UD}}                  & \multicolumn{3}{c}{\textbf{CTB6}}                \\ \cmidrule(l){2-13} 
                                        & $\mathcal{K}$=250          & $\mathcal{K}$=500          & $\mathcal{K}$=1000          & $\mathcal{K}$=250          & $\mathcal{K}$=500          & $\mathcal{K}$=1000          & $\mathcal{K}$=250          & $\mathcal{K}$=500          & $\mathcal{K}$=1000          & $\mathcal{K}$=250          & $\mathcal{K}$=500          & $\mathcal{K}$=1000          \\ \midrule
InstructUIE                             & 71.60          & 82.04          & 87.91          & 60.90          & 64.83          & 68.92          & 61.70          & 62.98          & 64.05          & 53.96          & 54.47          & 55.13          \\
GoLLIE                                  & 84.44          & 86.92          & 89.66          & 67.44          & 69.76          & 71.70          & 83.74          & 87.30          & \underline{89.54}    & \underline{81.35}    & \underline{86.21}    & \underline{88.42}    \\
GNER                                    & 83.91          & 86.13          & 87.45          & 71.78          & 71.99          & 72.34          & 80.10          & 82.35          & 86.99          & 75.09          & 79.00          & 83.73          \\ \midrule
\textbf{DIRECT(LLaMA-3.1-8B-Instruct)}   & \textbf{85.34}    & \underline{87.05}    & \textbf{90.10} & \underline{85.15}    & \underline{86.65}    & \textbf{88.03} & \underline{83.87}    & \underline{87.32}    & 88.19          & 81.30          & 85.61          & 87.66          \\
\textbf{DIRECT(GLM4-9B-Chat)}            & \textbf{85.34} & \textbf{87.17} & \underline{89.96}    & \textbf{85.47} & \textbf{86.77} & \underline{87.66}    & \textbf{84.48} & \textbf{87.44} & \textbf{89.76} & \textbf{81.72} & \textbf{87.14} & \textbf{89.59} \\ \bottomrule
\end{tabular}
}
\end{subtable}
\end{table*}

\subsection{Precision-Oriented Model Optimization}

To enhance the ability of LLMs to capture the characteristics of sequence labeling tasks and generate outputs that more closely align with the expected results, we apply DPO training after SFT, guided by a refined data selection strategy. Specifically, the SFT-trained LLM is first used to generate candidate answers for each input $X$, from which we construct preference pairs $(X, O^{+}, O^{-})$, where $O^{+}$ denotes the preferred answer and $O^{-}$ denotes the less-preferred one. To maximize entropy gain during training, we adopt an offline sampling strategy that integrates the BLEU score \cite{papineni2002bleu} with an F1-based template matching method for entity extraction. Among the sampled responses, those with the highest BLEU score but the lowest F1 score are selected as less-preferred responses, while the ground-truth answer is designated as the preferred response.

After constructing the preference pairs $(X, O^{+}, O^{-})$, both the reference model $\pi_{\mathrm{ref}}$ and the trainable discriminator model $\pi_{\theta}$ are initialized with parameters obtained from SFT. The parameters of the reference model remain fixed, while only the discriminator is updated during the DPO stage. For each preference triplet, we compute the log-probability differences under the discriminator and the reference model, respectively:
\begin{align}
\Delta_{\theta}
&= \log \pi_{\theta}(O^{+}\mid X) - \log \pi_{\theta}(O^{-}\mid X)
\label{eq:deltatheta}, \\
\Delta_{\mathrm{ref}}
&= \log \pi_{\mathrm{ref}}(O^{+}\mid X) - \log \pi_{\mathrm{ref}}(O^{-}\mid X).
\label{eq:deltaref}
\end{align}
The objective of DPO is to maximize the log-probability gain of the preferred output under the discriminator model relative to the reference model, which is typically achieved by minimizing a binary classification likelihood loss:
\begin{equation}\label{eq:dpo_pair_loss}
\mathcal{L}_{\mathrm{DPO}} =
- \log \sigma\!\Big(\beta\big(\Delta_{\theta} - \Delta_{\mathrm{ref}}\big)\Big),
\end{equation}
where $\sigma$ denotes the sigmoid function, and $\beta$ is a scaling hyperparameter.

\subsection{Inference Rectification}
During the inference stage, we incorporate a precise intervention mechanism into the generation process of LLMs. This mechanism ensures that the output strictly adheres to a predefined format while also improving inference efficiency through the integration of Key-Value (KV) Cache and prefilling techniques. Specifically, we first tokenize the complete set of possible labels $L = \{ l_1, l_2, \cdots, l_K \}$ and standardize them into token sequences of fixed length $T$, denoted as $S = \{ s_1, s_2, \cdots, s_K \}$. The core of our method is an iterative procedure that predicts a label for each input word $x_i$ by efficiently managing the context’s KV Cache. The process begins with context prefilling. Before predicting the label for a given word $x_i$, we process the preceding sequence to obtain the context $C_{t-2}$. A prefilling operation is then performed to compute and store the KV Cache for all tokens in $C_{t-2} = \left\{ h_1, h_2 \cdots, h_{t-2} \right\}$ in a single pass, thereby avoiding redundant attention computations for this known prefix in subsequent steps.

Next, we construct the initial context for predicting the token of $s_i$ as $C_{i} = C_{t-2} \oplus h_{\text{⟨LBR⟩}} \oplus h_{[t:t+T]}$, where $\oplus$ denotes sequence concatenation and ⟨LBR⟩ represents the token sequence for the left parenthesis “(”. During computation, it is only necessary to incrementally compute the key-values for ⟨LBR⟩ and $s_i$ and append them to the existing KV Cache of $C_{t-2}$. To generate the label token $u_{i,j} \in s_i$, the model produces the label of length $T$ in an auto-regressive, token-by-token manner. At each step, the model takes as input the KV Cache corresponding to the context $C_{t-1+j-1}$ and computes the probability distribution for the next token, which corresponds to the conditional probability $p_{\theta}(u_{i,j} \mid C_{t-1+j-1})$.
To ensure the validity of the generated sequence, we impose strict constraints on the decoding space. The model is restricted to selecting tokens from a predefined candidate set $V = \{ u_1, u_2,\cdots,u_m  \}$, where $u_{i,j} \in s_i \subseteq V_t$. The final prediction is obtained by choosing the token with the highest probability:
\begin{equation}
\label{eq:selection} 
u_{i,j} = \arg\max_{u_{i,j}\in V} p_{\theta} (u_{i,j} \mid C_{t-1+j-1} ).
\end{equation}
After selecting the tokens $s_i$, we update the context to $C_i = C_{t+T}$ and incrementally update its corresponding key-values. At this stage, both the subsequent right parenthesis ⟨RBR⟩ and the next input word $x_{i+1}$ are already determined. To prepare for predicting the label of $x_{i+1}$, we continue the prefilling process for the subsequent known content based on the current KV Cache. In this way, the reuse of the existing KV Cache enables seamless iteration while avoiding redundant computations for known sequences. By combining this constrained generation strategy with KV Cache prefilling, we not only enforce the generation of syntactically valid labels but also maximize computational efficiency. Multiple attention operations are consolidated into a single prefill pass for known prefixes, which yields a substantial improvement in inference speed while preserving output quality.

\section{Experiments}
\label{sec:experiments}

\subsection{Experimental Setup}

\textbf{Datasets} We evaluate the performance of our model on six NER datasets and two POS datasets. The NER datasets include four Chinese datasets Weibo \cite{peng2015named}, Youku \cite{jie2019better}, Taobao \cite{jie2019better}, and Resume \cite{zhang2018chinese}, as well as two English datasets MIT-Movie \cite{liu2013asgard} and CoNLL03 \cite{sang2003introduction}. The POS datasets include two Chinese datasets, UD \cite{nivre2016universal} and CTB6 \cite{xue2005penn}. To assess the model's performance in low-resource scenarios, we randomly sample a subset of $\mathcal{K}$ samples from the training set for training, respectively, where $\mathcal{K}$ is set to 250, 500, and 1000.

\textbf{Baselines} To evaluate our model's performance on low-resource sequence labeling tasks, we compare it with the following baselines. (1) \textbf{InstructUIE} \cite{wang2023instructuie}, which is based on Flan-T5-xxl. (2) \textbf{GoLLIE} \cite{sainz2023gollie}, which is based on Code-LLaMA-7B. (3) \textbf{GNER} \cite{ding2024rethinking}, which is based on LLaMA-7B.

\textbf{Training Details} To better evaluate the effectiveness of our approach, we employ two LLMs, LLaMA-3.1-8B-Instruct \cite{dubey2024llama} and GLM-4-9B-Chat \cite{glm2024chatglm}, as our backbones for comparison. For DPO training, we set the hyperparameter $\beta$ to 0.1. In addition, since the Flan-T5 model used in InstructUIE is limited in its ability to encode Chinese, we adopt the mT5-xxl \cite{xue2020mt5} model from the T5 family for experiments on Chinese datasets, while retaining Flan-T5-xxl for English datasets. 

\subsection{Main Results}
We evaluate the performance of different methods using the F1-score, and the results are presented in Table~\ref{tab1}. The experiments show that when DIRECT adopts GLM4-9B-Chat as the backbone model, it consistently achieves the best performance across virtually all NER and POS datasets under different values of $\mathcal{K}$. When DIRECT employs LLaMA-3.1-8B-Instruct as the backbone, it generally attains the second-best results across datasets. Specifically, under different $\mathcal{K}$ settings, the average F1-score of DIRECT improves by 0.98\%, 5.22\%, 4.25\%, 2.55\%, 0.53\%, 14.72\%, 0.37\%, and 0.82\% on the Youku, Taobao, Weibo, Resume, CoNLL03, MIT-Movie, UD, and CTB6 datasets, respectively. Notably, on the MIT-Movie dataset, compared to other baseline models, our method achieves substantial improvements, with F1-scores increased by 13.69\%, 14.78\%, and 15.69\% under $\mathcal{K}=250$, $\mathcal{K}=500$, and $\mathcal{K}=1000$, respectively. We attribute these performance gains to the incorporation of preference pairs for DPO optimization during training, as well as controlled interventions in the generation process during inference, which enable the model outputs to better align with the expected results.

\begin{table}[]
\caption{Ablation study on different backbone LLMs with $\mathcal{K}=1000$. The best results are highlighted in \textbf{bold}.}\label{tab2}
\centering
\begin{subtable}{\columnwidth}
\resizebox{\columnwidth}{!}{%
\begin{tabular}{@{}lcccc@{}}
\toprule
\multicolumn{1}{c}{\textbf{}}          & \textbf{Youku} & \textbf{Taobao} & \textbf{Weibo} & \textbf{Resume} \\ \midrule
\textbf{DIRECT(LLaMA-3.1-8B-Instruct)} & \textbf{75.72} & \textbf{78.68}  & \textbf{70.52} & \textbf{93.13}  \\
w/o DPO                                & 75.37          & 78.34           & 68.66          & 94.94           \\
w/ SFT                                 & 74.50          & 67.68           & 65.51          & 92.49           \\ \midrule
\textbf{DIRECT(GLM4-9B-Chat)}          & \textbf{79.31} & \textbf{78.58}  & \textbf{70.31} & \textbf{95.94}  \\
w/o DPO                                & 78.72          & 78.27           & 68.94          & 95.78           \\
w/ SFT                                 & 78.48          & 70.52           & 60.82          & 92.56           \\ \bottomrule
\end{tabular}
}
\end{subtable}

\vspace{0.2em} % 控制上下间距

\begin{subtable}{\columnwidth}
\resizebox{\columnwidth}{!}{%
\begin{tabular}{@{}lcccc@{}}
\toprule
\multicolumn{1}{c}{\textbf{}}          & \textbf{CoNLL03} & \textbf{MIT-Movie} & \textbf{UD}    & \textbf{CTB6}  \\ \midrule
\textbf{DIRECT(LLaMA-3.1-8B-Instruct)} & \textbf{90.10}   & \textbf{88.03}     & \textbf{88.19} & \textbf{87.66} \\
w/o DPO                                & 89.97            & 87.80              & 88.08          & 87.06          \\
w/ SFT                                 & 88.97            & 67.82              & 86.55          & 81.86          \\ \midrule
\textbf{DIRECT(GLM4-9B-Chat)}          & \textbf{89.96}   & \textbf{87.66}     & \textbf{89.76} & \textbf{89.59} \\
w/o DPO                                & 89.59            & 87.59              & 89.43          & 89.14          \\
w/ SFT                                 & 88.76            & 67.92              & 86.40          & 81.35          \\ \bottomrule
\end{tabular}
}
\end{subtable}
\end{table}

\subsection{Ablation Study}

To further validate the effectiveness of each component and strategy in DIRECT, we conducted ablation experiments under the setting of $\mathcal{K}=1000$ across all datasets and backbone LLMs, with the results summarized in Table~\ref{tab2}. (1) ``w/o DPO'' denotes the removal of the DPO strategy during training. The results show that this modification consistently leads to performance degradation across all datasets and backbone LLMs. This is mainly because, without the DPO strategy, the model cannot learn effective contrastive signals from preference pairs, making it difficult to establish decision boundaries that are well aligned with the requirements of sequence labeling tasks, ultimately resulting in degraded performance. (2) ``w/ SFT'' indicates performing only SFT on the backbone model without incorporating additional DPO training or inference intervention. The results reveal that this modification leads to further performance decline. Notably, when the backbone LLM is LLaMA-3.1-8B-Instruct, the F1-scores of DIRECT on the Taobao and MIT-Movie datasets drop by 11.00\% and 20.21\%, respectively. These findings demonstrate that, through DPO training and inference correction, our method achieves significant performance improvements over the original backbone LLMs. DPO training enables the model to better align with preference signals, while inference correction effectively reduces redundancy and uncertainty in the generation process, thereby delivering superior generalization and alignment performance across diverse datasets and scenarios.

\begin{figure}[ht]%
\centering
\includegraphics[width=\columnwidth]{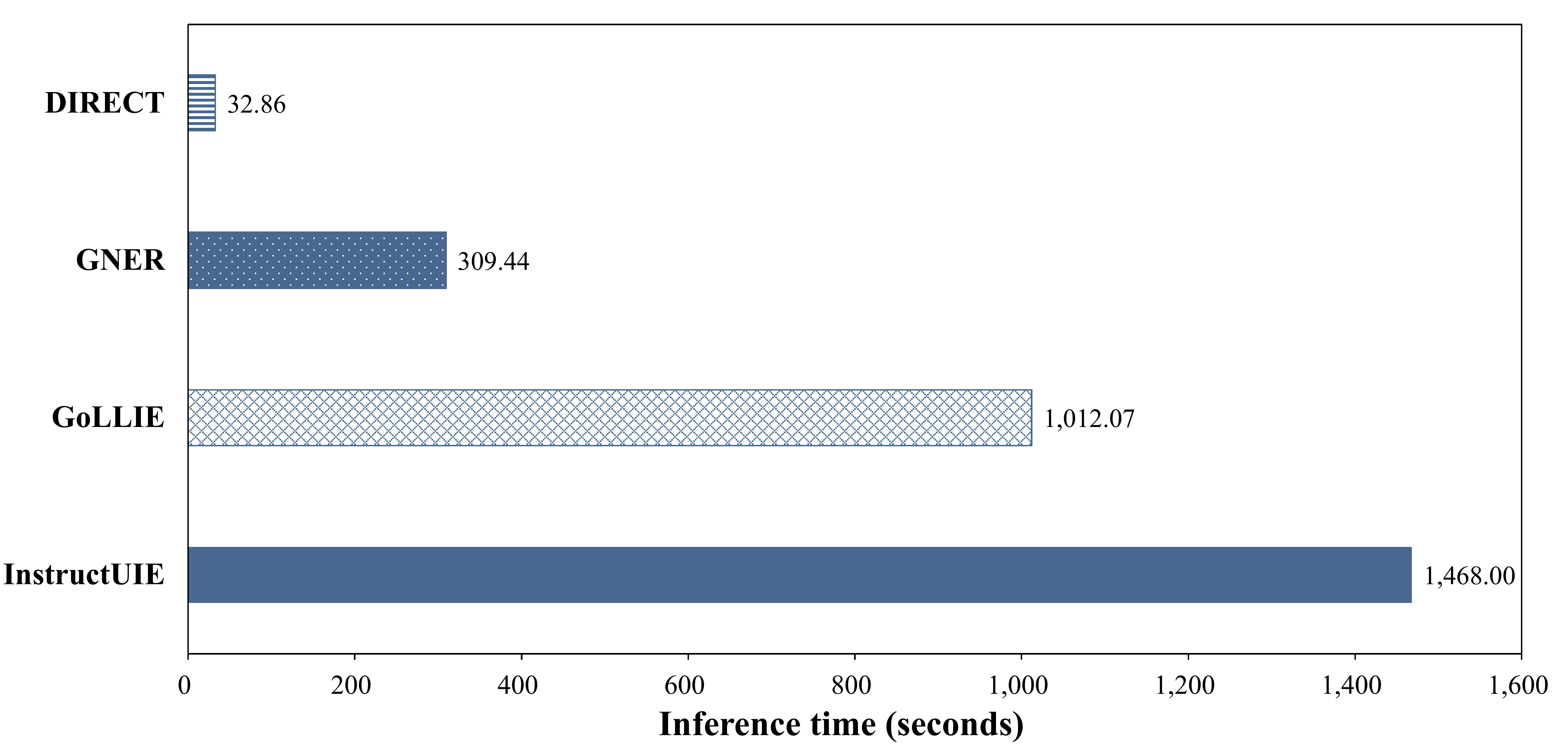}
\caption{Inference efficiency comparison of different methods, where DIRECT adopts LLaMA-3.1-8B-Instruct as the backbone.}\label{fig2}
\end{figure}

\subsection{Inference Time Analysis}
To validate the efficiency of DIRECT in the inference stage, we compare its inference speed against three representative baseline models, with the results shown in Figure~\ref{fig2}. To ensure fairness and maintain consistency with the settings of other methods, we fix the batch size to 1 and conduct experiments on a single NVIDIA L40 GPU. The experimental data are sampled from 10 sentences in the CTB6 dataset, with an average length of 192 tokens. The results demonstrate that DIRECT significantly outperforms all baselines in inference speed, requiring only 32.86 s to complete inference on all sentences, which is about one-tenth of the strong baseline GNER. This efficiency gain primarily stems from the template-filling mechanism adopted during inference, where the LLMs only need to generate the predicted label tokens, while the remaining parts are automatically filled by the template. This reduces redundant generation and substantially accelerates the inference process. These findings highlight that DIRECT not only achieves strong performance but also demonstrates superior practical deployability.

\begin{figure}[ht]%
\centering
\includegraphics[width=\columnwidth]{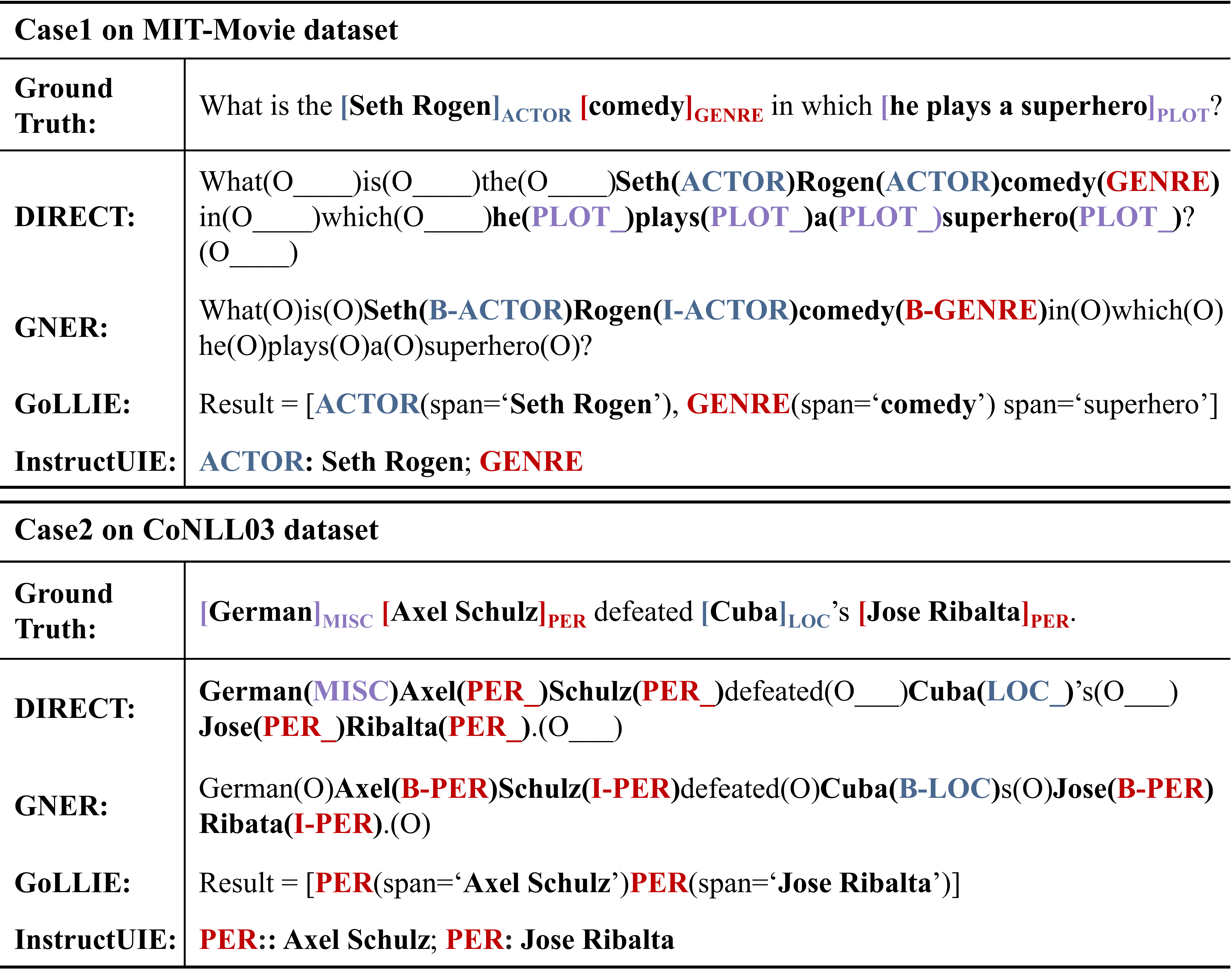}
\caption{Case study with DIRECT and baselines, where DIRECT adopts LLaMA-3.1-8B-Instruct as the backbone.}\label{fig3}
\end{figure}

\subsection{Case Study}
To provide a more intuitive comparison of the advantages of DIRECT over other methods, we conduct case studies in Figure~\ref{fig3}. In case 1 and 2, the other three baseline models all fail to correctly identify the ``PLOT'' entity or the ``MISC'' entity. At the same time, these baselines generally suffer from issues of missing outputs or redundant outputs. For example, in case 1, the output of GNER omits the word ``the'' and its corresponding label, as well as the label for the final word ``?''; GoLLIE produces an output where ``span=`superhero' '' does not conform to its predefined formatting rules; and InstructUIE fails to provide the corresponding entity content after the ``GENRE'' label. In contrast, DIRECT not only identifies and classifies all entities completely and accurately in both cases but also strictly adheres to the expected output format, thereby demonstrating stronger robustness and controllability.

\section{Conclusion}
In this paper, we propose DIRECT, a DPO-based inference rectification method. To enhance the alignment capability of LLMs in domain adaptation, we perform DPO optimization with preference pairs during training and intervene in the generation process during inference, ensuring that the outputs of LLMs are better aligned with human preferences and the requirements of sequence labeling tasks. Furthermore, to improve inference efficiency, we introduce a template-filling mechanism that substantially reduces redundancy in generation. Experimental results demonstrate that DIRECT achieves significant improvements in both performance and efficiency compared with previous methods.

%\vfill
%\pagebreak

\vfill\pagebreak

% \section{REFERENCES}
% \label{sec:refs}

% List and number all bibliographical references at the end of the
% paper. The references can be numbered in alphabetic order or in
% order of appearance in the document. When referring to them in
% the text, type the corresponding reference number in square
% brackets as shown at the end of this sentence \cite{C2}. An
% additional final page (the fifth page, in most cases) is
% allowed, but must contain only references to the prior
% literature.

% Please follow the IEEE Citation Guidelines, \url{https://ieee-dataport.org/sites/default/files/analysis/27/IEEE\%20Citation\%20Guidelines.pdf} for formatting of references.


{
\begin{spacing}{0.9}
\bibliographystyle{IEEEbib}
\bibliography{strings,refs}
\end{spacing}
}

\end{document}